\documentclass[conference]{IEEEtran}
\IEEEoverridecommandlockouts

\usepackage{cite}
\usepackage{amsmath,amssymb,amsfonts}
\usepackage{algorithmic}
\usepackage{graphicx}
\usepackage{textcomp}
\usepackage{xcolor}
\usepackage{listings}
\usepackage{scrextend}
\usepackage{hyperref}

\newcommand{\yantian}[1]{{\color{black} #1\color{black}}}
\newcommand{\snehesh}[1]{{\color{black} #1\color{black}}}

\def\BibTeX{{\rm B\kern-.05em{\sc i\kern-.025em b}\kern-.08em
    T\kern-.1667em\lower.7ex\hbox{E}\kern-.125emX}}
\begin{document}

\title{NatSGLD: A Dataset with \underline{S}peech, \underline{G}esture, \underline{L}ogic, and \underline{D}emonstration for Robot Learning in \underline{Nat}ural Human-Robot Interaction$^\textsuperscript{\textdagger}$\thanks{$\textsuperscript{\textdagger}$ This work has been accepted for presentation at the 20th edition of the ACM/IEEE International Conference on Human-Robot Interaction.}\\
}

\author{\IEEEauthorblockN{
Snehesh Shrestha, Yantian Zha$^*$, Saketh Banagiri, Ge Gao$^*$, Yiannis Aloimonos$^*$, Cornelia Fermüller$^*$}
\IEEEauthorblockA{
\textit{University of Maryland}\\
College Park, USA \\
\{snehesh,ytzha,sbngr,gegao,jyaloimo,fer\}@umd.edu}\thanks{$^*$ The authors equally advised the research}
}

\maketitle

\begin{abstract}
Recent advances in multimodal Human-Robot Interaction (HRI) datasets emphasize the integration of speech and gestures, allowing robots to absorb explicit knowledge and tacit understanding. However, existing datasets primarily focus on elementary tasks like object pointing and pushing, limiting their applicability to complex domains. They prioritize simpler human command data but place less emphasis on training robots to correctly interpret tasks and respond appropriately. To address these gaps, we present the NatSGLD dataset, which was collected using a Wizard of Oz (WoZ) method, where participants interacted with a robot they believed to be autonomous. NatSGLD records humans' multimodal commands (speech and gestures), each paired with a demonstration trajectory and a Linear Temporal Logic (LTL) formula that provides a ground-truth interpretation of the commanded tasks. This dataset serves as a foundational resource for research at the intersection of HRI and machine learning. By providing multimodal inputs and detailed annotations, NatSGLD enables exploration in areas such as multimodal instruction following, plan recognition, and human-advisable reinforcement learning from demonstrations. We release the dataset and code under the MIT License at \url{https://www.snehesh.com/natsgld/} to support future HRI research.
\end{abstract}

\vspace{-5mm}
\section{Study Overview}
Human communication often involves the concurrent use of language and gestures, as noted by Tomasello (\cite{tomasello2010origins}, p. 230). Language-based communication conveys explicit knowledge, gestures excel at expressing tacit, context-dependent information, and together they allow people to express with efficient clarity. For tasks involving both explicit and \yantian{tacit knowledge,} such as food preparation, cooking, and cleaning, this kind of hybrid communication approach becomes essential, especially in natural human-robot interactions\footnote{Natural here means humans interact with robots as if they were interacting with other humans.}.

\begin{figure*}[ht]
\centering
\includegraphics[width=1.\textwidth]{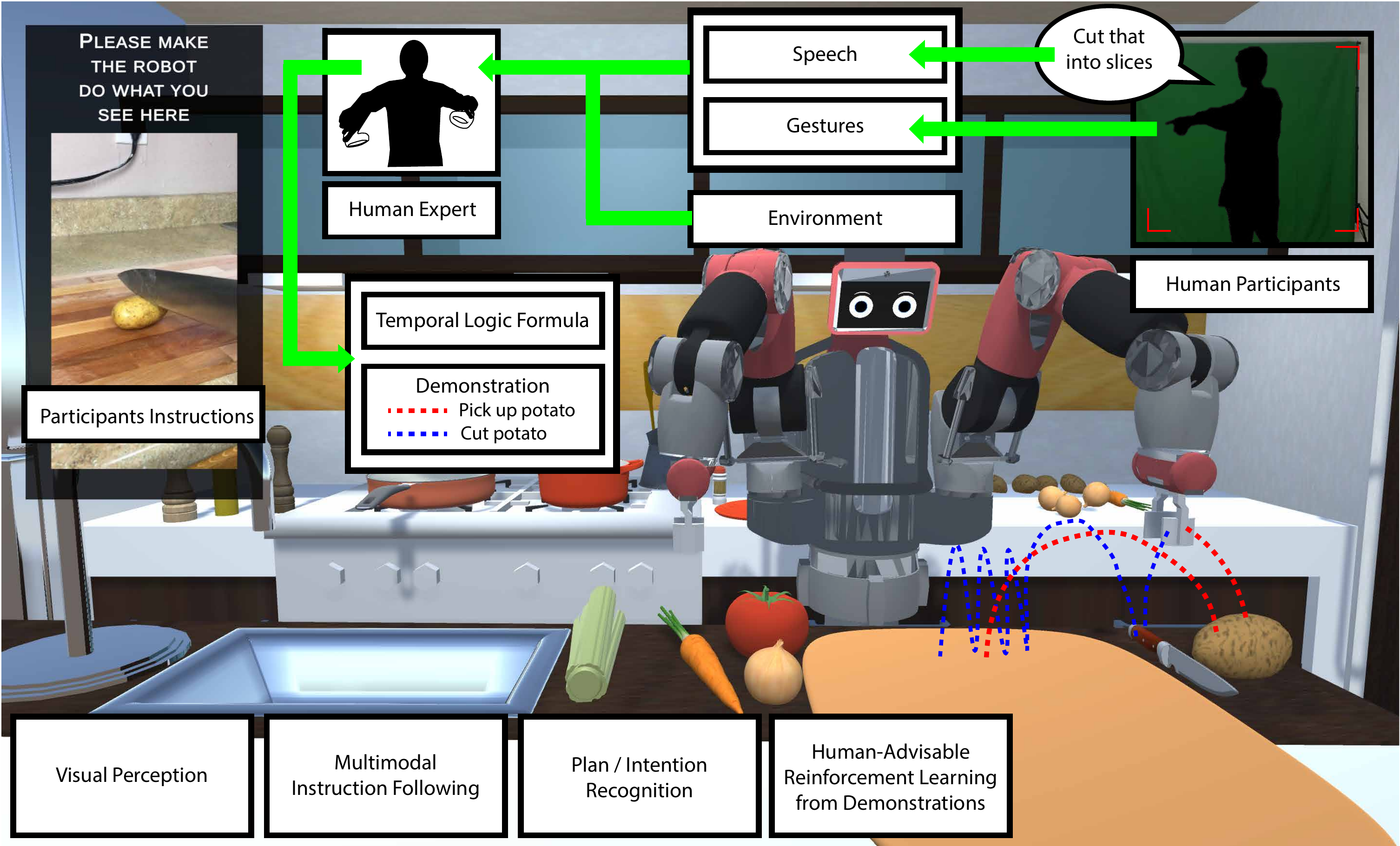}
\vspace{-2mm}\caption{NatSGLD contains speech, gestures, temporal logic annotations, and demonstration trajectories for  everyday food preparing, cooking, and cleaning tasks. The NatSGLD dataset enables the learning of complex human-robot interaction tasks due to the rich interaction modalities and strong supervising signals at both trajectory-level (demonstrations) and symbolic-level (ground-truth temporal logic formulas that match humans' expectation on the robot).}
\label{fig:intro}\vspace{-6mm}
\end{figure*}

Enabling robots to collaborate with humans in a natural way can significantly reduce cognitive load for humans. For example, as shown in Fig. \ref{fig:intro}, a human could simply say, ``Cut that into pieces", while pointing at a potato. The gesture conveys tacit knowledge, such as the location of the potato, in a way that is intuitive for both the human and the robot. Similarly, a command like ``Please stir that" accompanied by an enactment of stirring, communicates not only what to stir but also how -- slowly, gently, rapidly, or in specific directions.

In these natural interactions, humans often focus on the critical subtasks and assume the robot possesses the commonsense knowledge to handle the rest. For instance, a human might assume the robot knows it needs to pick up a knife before cutting the potato. \yantian{This requires the robot to predict what the human expects it to do based on multimodal inputs (speech and gestures), the current observation of the environment, and translating these into symbolic task representations, such as temporal logical formulas. This alignment with human expectations allows the robot to generate motion trajectories that effectively carry out the task.}

However, most existing Human-Robot Interaction (HRI) datasets either focus exclusively on speech (e.g., Google Home, Amazon Alexa) \cite{novoa2017multichannel, james2018open, vasudevan2018object, narayan2019collaborative, padmakumar2022teach}, or solely on gestures \cite{pisharady2015recent, shukla2016multi, mazhar2018towards, chen2018wristcam, chang2019improved, gomez2019caddy, neto2019gesture, nuzzi2021hands, de2021introducing}. Some combine speech and gestures but are limited to perception tasks such as object recognition or manipulation \cite{matuszek2014learning, rodomagoulakis2016multimedia, azagra2017multimodal, chen2022real}. These datasets do not fully capture the complexity of natural, multimodal communication in real-world tasks.

\begin{figure*}[ht]
\centering
\includegraphics[width=1\textwidth]{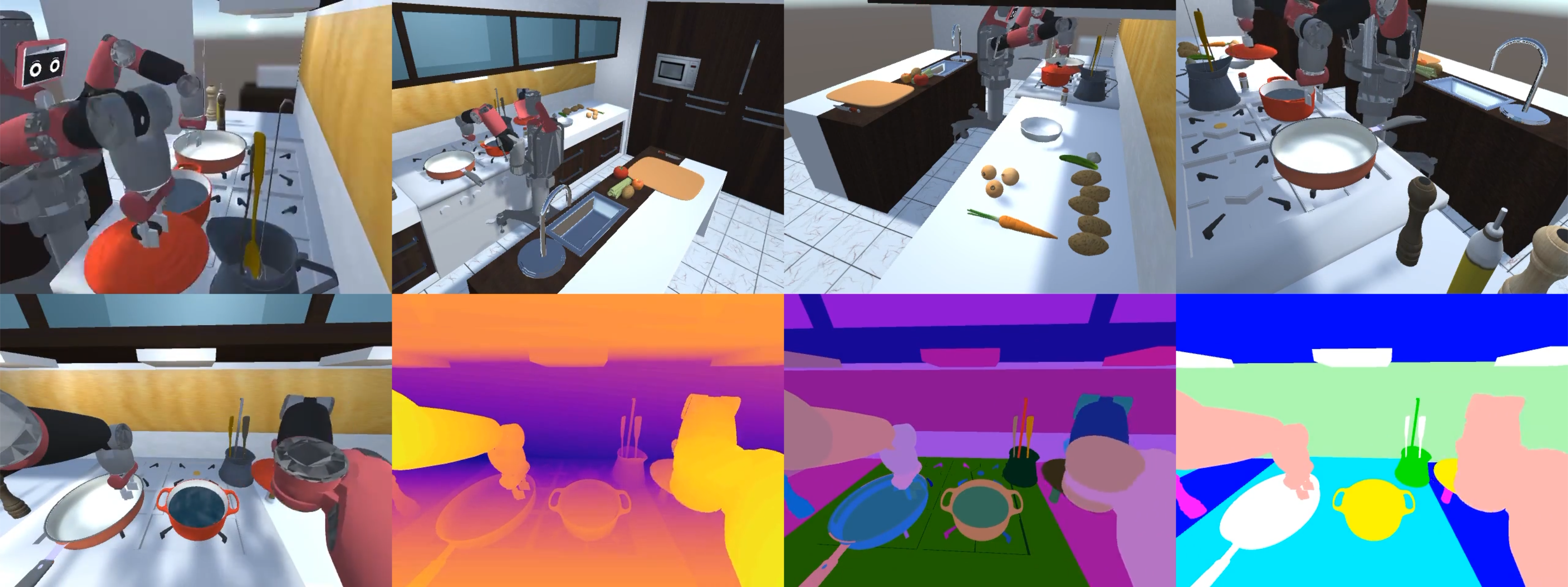}
\vspace{-2mm}\caption{Illustration of Our Simulator (Baxter Opens Pot Lid): Our simulator provides a comprehensive, multi-view perspective of the robot and its environment, captured through both static and mobile cameras. The top row includes three camera angles: the human's first-person view, an overhead shot (top left), and two views from the kitchen counter (top right and left). The bottom row presents the robot's egocentric view, featuring multiple sensor outputs: RGB, depth images, distinct object segmentation map, and category-based semantic segmentation map. These varied perspectives enable the robot to interpret and perform tasks via human speech and gesture commands, as well as autonomously evaluate its surroundings and the state of objects.}
\label{fig:natsgd_multi_cam_views}
\vspace{-5mm}
\end{figure*}

To address this gap, we introduce the NatSGLD dataset alongside a simulator built on the Unity game engine (described in Fig. \ref{fig:intro} and Fig. \ref{fig:natsgd_multi_cam_views}). This dataset enables our simulated robot to understand natural human speech and gesture commands for complex tasks such as food processing and cooking. We annotated these multimodal commands with expert-teleoperated robot demonstrations within the simulator and ground-truth understanding of participant-commanded tasks as Linear Temporal Logic (LTL) formulas\footnote{For a brief introduction to LTL, visit \url{https://www.cwblogs.com/posts/linear-temporal-logic/}. LTL provides a flexible, compact method for describing complex tasks with intricate temporal properties. See Sec. \ref{sec:ltl} for details.}. 

We employed a Wizard of Oz (WoZ) method \cite{Dahlback1993-el}, where participants interacted with a robot they believed to be autonomous. NatSGLD includes 1,143 multimodal commands from 18 participants across 11 activities, 20 objects, and 16 object states. The dataset provides a rich collection of data, including \yantian{robot states, participants' speech transcripts, gestures, gazes, and expert annotations of tasks and robot trajectories, while offering multiple perspectives such as diverse camera views, depth images, and semantic segmentation (Fig. \ref{fig:natsgd_multi_cam_views}).}

\yantian{To the best of our knowledge, NatSGLD is the first HRI dataset to integrate speech, gestures, annotated logical task representations, and expert demonstrations. This unique combination enables robots to handle complex, natural interactions and fosters research in multimodal HRI. By adapting state-of-the-art reinforcement learning, instruction following, and plan recognition techniques such as \cite{zha2018recognizing,zhuo2020discovering,ha2023scaling,kambhampati2022symbols,ding2023task,venkataraman2024real,qiu2024eillm,cui2021empathic,zhang2021recent,chen2024reltrans,nair2018overcoming,mees2022calvin,zha2024learning,shervedani2023multimodal,lin2023gesture,wang2024language,lu2022system,deichler2023learning,shao2021concept2robot,xiao2022robotic,goyal2021pixl2r,zhou2021inverse,li2023proactive,lin2022inferring,du2024learning,yan2021human,gross2024communicative,fiorini2021daily,hwang2022human,wang2021multimodal,zhang2024vision,ekrekli2023co,zamani2018learning,du2018online,cruz2018multi,trick2022interactive}, NatSGLD supports key applications, including multimodal instruction following, plan recognition, human-advisable reinforcement learning, and perception tasks like object, gesture, and activity recognition.}

\section{Methods}
\label{sec:methods}
\subsection{Simulation Development}
Current robots are unable to perform kitchen tasks at real-time speeds that humans naturally expect from one another. Interacting with slowly-moving robots can influence how human participants command them, affecting their choice of language and gestures. To address this, we developed a real-time photorealistic simulator for HRI research, built using Unity3D \cite{Unity_Technologies_undated-qy} and integrated with a customized ROS plugin \cite{Bischoff_rossharp}. Our simulator facilitates real-time execution of robotic tasks, thereby enhancing its applicability for future HRI studies.

\subsubsection{Simulator Design and Capabilities}
Our simulator is designed to handle intricate, multi-step tasks commonly encountered in real-world scenarios, such as cutting or pouring. For instance, during a task like cutting a tomato, the robot must locate and grasp the knife, stabilize itself, and perform a series of cuts. Although perfect simulation of such activities remains an open challenge, our simulator leverages state-of-the-art techniques to represent cutting by dynamically replacing object meshes with segmented parts once cut actions are initiated. Additionally, the simulator generates a variety of sensor data types essential for robot learning, including multi-camera RGB image views, depth images, semantic and instance segmentation maps (Fig. \ref{fig:natsgd_multi_cam_views}). These outputs, essential for training vision-based models, enhance the simulator's utility for learning complex tasks through multimodal inputs.

\subsubsection{Natural Interaction with Participants}


The simulator ensures seamless real-time interaction between participants and the robot. It operates on a high-performance computer (Intel i7, 16GB RAM) connected to a 55" display, providing smooth processing and rendering. A live camera feed of participants is overlaid on the screen, giving the impression that the robot can ``see" them, enhancing immersion. The robot maintains eye contact with humans when listening to them and focuses on objects while performing actions. The simulator captures both human and robot perspectives, providing comprehensive data on object states, robot trajectories, and task progression.

\vspace{-1mm}
\subsection{Data Collection}
\yantian{The study, approved by our university's Institutional Review Board, involved 18 participants (aged 18 to 31 years, Mean: 21, SD: 4) with equal gender distribution (9 male, 9 female) and a mix of technical (44\%) and non-technical (56\%) backgrounds. None had prior experience interacting with robots.} \snehesh{See Appendix \ref{appendix:participants} for details.} 

\subsubsection{Wizard-of-Oz Experiment Design}
\begin{figure}[ht]
\centering
\includegraphics[width=0.92\columnwidth]{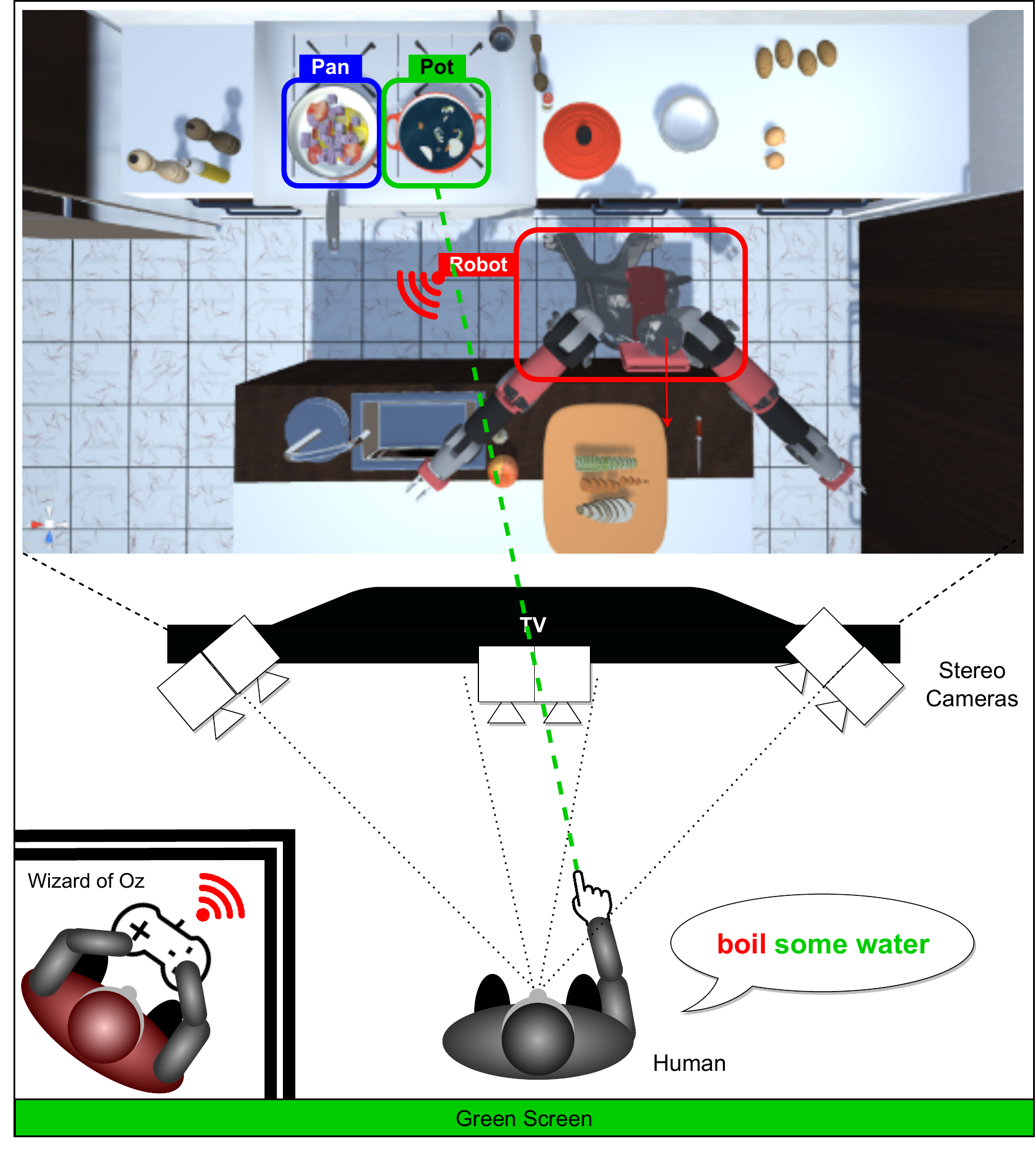}
\vspace{-3mm}\caption{NatSGLD experiment setup (top view). The participant points at the pot on the right stove and asks the robot to boil water, implying the right burner should be turned on. The bottom-left corner shows the location of the hidden Wizard of Oz, who observes the participant and controls the robot.}
\label{fig:NatComm_Dataset_TopAndFrontView_PourVeges}\vspace{-7mm}
\end{figure}

To ensure the naturalness of human behavior in our study, we implemented a Wizard-of-Oz (WoZ) experimental design \cite{Dahlback1993-el}, where participants were led to believe they were interacting with a fully autonomous and capable robot. To enhance this illusion, participants first watched a video demonstrating the robot performing complex tasks autonomously. During the practice session, they issued commands to the robot, which it seamlessly executed without errors, reinforcing the perception of its autonomy. Unbeknownst to them, a concealed operator controlled the robot’s actions in real time (see Fig. \ref{fig:NatComm_Dataset_TopAndFrontView_PourVeges}), ensuring precision while maintaining the illusion of autonomy. This approach fostered spontaneous and authentic interactions between participants and the robot. See \textit{Appendix \ref{appendix:woz}} for details.

\subsubsection{Linear Temporal Logic (LTL)}\label{sec:ltl}
Challenging tasks like food preparation and cooking, commanded by participants, often involve both sequential and concurrent actions. LTL formally captures these temporal relations \cite{kesten1998algorithmic}, which is crucial for tasks like ``Chop the vegetables while preheating the oven". Additionally, LTL offers a compact, flexible task representation that can be converted into automata \cite{gastin2001fast, babiak2012ltl}, allowing the extraction of symbolic plans. Thus, we use LTL to annotate the ground-truth interpretation of participants' multimodal commands for training robots. In the example from Fig. \ref{fig:intro}, the derived LTL formula is:


\begin{addmargin}[2em]{0em} 
\begin{scriptsize}
\texttt{X (G (C\_Potato U Potato\_FarFrom\_CT) \& G F (Potato\_OnTopOf\_CB \& Potato\_CloseTo\_CB) \& X ( G (C\_Knife U Knife) \& G (C\_Knife U Knife\_FarFrom\_CT) \& G (C\_Knife U Knife\_OnTopOf\_Potato) \& G ((C\_Potato \& Knife\_CloseTo\_Potato) U Potato\_Pieces)))}
\end{scriptsize}
\end{addmargin} Here, ``X," ``F", ``U", and ``G" are LTL operators representing ``neXt", ``Finally", ``Until", and ``Globally (Always)" respectively. Additionally, ``C\_Potato", ``Potato\_FarFrom\_CT", ``Potato\_CloseTo\_CB", and ``Potato\_Pieces" are grounded predicates signifying symbolic relations among objects, where ``CB" stands for ``Cutting Board," ``CT" for ``Countertop," and ``C" indicates ``Holds". 

\subsubsection{Robot Demonstrations}
Alongside LTL, we collected expert-teleoperated demonstrations to train robots at the control level by capturing detailed sensor data and joint controls. This is crucial for training and evaluating robot decision-making models. The combination of LTL and demonstrations enables the dataset to assess task understanding and motion execution across both simple and complex scenarios. To collect demonstrations, an expert teleoperated the robot head and end-effector, and the detailed joint angles were generated using real-time inverse kinematics (IK \cite{Starke2017-cx, Starke2018-pb}). All the sensor data and robot joint control outputs at each step were recorded.

\subsubsection{Post-processing and Annotation}
Participants' speech was transcribed manually by the researchers and encoded using the Glove model \cite{pennington2014glove} and BART \cite{lewis2019bart} to extract embeddings, while gestures were captured as a sequence of body poses using pose estimation algorithm OpenPose \cite{james2018open}. Each multimodal command was annotated with a human expert in  Linear Temporal Logic (LTL), reflecting the task's temporal structure and the subtasks inferred by the human based on both speech, gestures, and context. See \textit{Appendix \ref{appendix:post_processing_sync_cam_calib}} for details.

\subsubsection{Ethical and Bias Considerations}
To ensure fairness and minimize biases, we accounted for participant demographics, including gender, age, and personality traits. Data on participants' experiences, such as workload assessments via NASA-TLX \cite{hart2006nasa}, were also collected. These factors provide insights into how different participants perceive and interact with robots, which is critical for developing unbiased, generalizable human-robot interaction models.

\section{NatSGLD Dataset}
The NatSGLD dataset is organized into three main sections: the simulator, scripts, and data.

\subsection{Simulator}
The simulator is provided as a Unity 3D project, along with instructions for setting up third-party add-ons. It can be controlled using keyboard shortcuts, Python scripts, and ROS messages. Robot trajectories can also be replayed by specifying the trajectory records included in the dataset.

\subsection{Scripts}
There are four types of scripts: a) experiment execution scripts, b) data collection scripts, c) post-processing scripts (synchronization, extraction, parsing, compression), and d) analysis and annotation tools for replaying datasets and modifying annotations (e.g., speech, gestures, tasks, robot social behaviors like robot's eye contact, nodding, and confusion).

\subsection{Data}
\begin{itemize}
    \item \textbf{Metadata (.csv)}: 
    The metadata, stored as a CSV file, serves as the primary database. Each record corresponds to a command issued by participants, identified by a unique global record number (DBSN) and participant ID (PID). The sequence number of commands within a session is recorded as SSN. The onset (start time, ST) and offset (end time, ET) of each command are provided in milliseconds from the start of the experiment, and the presence of speech or gestures is indicated by a boolean value (1 for presence, 0 for absence).
    \item \textbf{Videos (.mp4)}: 
    Organized into subfolders by participant (e.g., P41, P45, P50), each folder contains various views:
        \begin{itemize}
            \item Participant View: Audio at 44kHz and video at 30fps
            \item Experiment View: Video at 30fps
            \item Multicam Room View: Video at 30fps
            \item Robot's First-Person (Ego) View: Video at 30fps, including RGB and depth images, instance segmentation maps (e.g., Onion1, Knife), and category segmentation maps (e.g., food, utensils, robot)
        \end{itemize}
                
    \item \textbf{Events (.dat)}: 
    Events refer to key activities during the experiments, including participant speech (transcribed as text), human gesture videos, tasks or subtasks (see \textit{Appendix \ref{appendix:subtasks}}), object state changes (e.g., object grabbed, cut; see \textit{Appendix \ref{appendix:objects}}), and robot actions (tasks or subtasks performed). All events were annotated using FEVA \cite{shrestha2023feva} and stored in .dat format, following a JSON structure. Each event is identified by a unique 16-character alphanumeric UUID (annotation ID), with start and end times in milliseconds and annotation text. Robot actions can involve task performance or social interactions.
    
    \item \textbf{States (.npz)}:
    State data is stored as compressed numpy files, with PID as the primary key and DBSN as the secondary key. The states include robot and object data. Robot states capture location (orientation and translation), joint states (single angle for the head, 7 angles for each arm in radians), and 2D bounding boxes of objects in the robot's ego-view (object ID, starting x and y coordinates, width, and height in pixels). Object states are one-hot encoded for objects whose state has changed, with detailed descriptions available in the events .dat file.
    
    \item \textbf{Features (.npz)}: 
    Features are stored as compressed numpy files, comprising embeddings and annotations, with PID as the primary key and DBSN as the secondary key. \yantian{Each speech command includes GloVe and BART embeddings. Gesture videos are processed using OpenPose's 25 keypoints, and human gaze data is extracted with L2CS-Net \cite{abdelrahman2023l2cs}, including face bounding boxes, gaze pitch/yaw angles, landmarks, and prediction confidence scores.} Tasks are represented in text as LTL formulas.

\end{itemize}

\section{Usage Notes}
\label{sec:usage_notes}
Access the NatSGLD simulator, dataset, and scripts at  \url{https://github.com/sneheshs/natsgld}. \yantian{The software was verified in Window 10, Mac OS Monterey, and 
Ubuntu 20.04.} The repository contains detailed documentation including dependencies, installations instructions, scripts for execution, data collection, post-processing, annotating, and data loading. \yantian{Appendix \ref{appendix:multi-modal-human-task-understanding} demonstrates an example of using NatSGLD to train robots for multimodal task understanding from speech and gestures.}

Given the potential for biases introduced by participant demographics, researchers should be mindful of ethical considerations when using the dataset. Participant diversity has been accounted for in the collection process, but users should still ensure that their models do not perpetuate unintended biases, particularly when generalizing across populations.

Researchers are encouraged to extend the dataset by collecting new demonstrations or by exploring additional domains where speech and gesture-based human-robot interaction is relevant. Scripts provided with the dataset allow for easy data collection, enabling researchers to contribute to the continuous improvement and expansion of NatSGLD.

\section{Acknowledgments}
\label{sec:acknowledgments}
The work was supported by NSF grant OISE 2020624. We thank the peers and faculty members from the University of Maryland for their feedback, especially Dr. Vibha Sazawal, Dr. Michelle Mazurek, Vaishnavi Patil, and Lindsay Little. We thank the contributions of the students listed in Appendix \ref{appendix:additional_acknowledgements}.

\bibliographystyle{ieeetr}
\bibliography{output}

\clearpage

\begin{appendices}

\section{Participants Details}
\label{appendix:participants}
\snehesh{During recruitment, we aimed for balanced demographics in gender, age, and personality traits. We initially screened for diversity in gender and age and adjusted recruitment to address any underrepresented groups. Although we didn’t screen for personality traits upfront, we collected self-reported personality data and reviewed it post-collection and validated that the participants had a mix of traits to avoid potential bias.}

\begin{figure}[h]
    \centering
    \includegraphics[width=\linewidth]{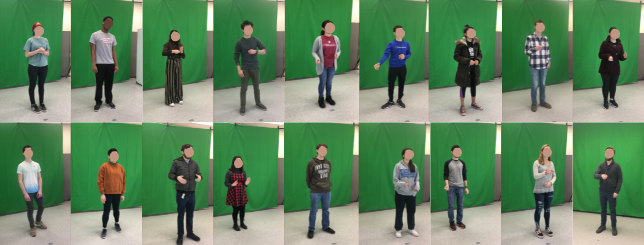}
    \caption{In this example, the participants are instructing the robot to cut onions where the participant's natural choice of communication is diverse.}
    \label{fig:appendix_participants}
\end{figure}

\section{Design Details of Wizard-of-oz (WoZ) Experiments}
\label{appendix:woz}

\subsection{Briefing and Debriefing}
\label{appendix:briefing_and_debriefing}
\yantian{To balance the spontaneity of natural behavior with the control of a lab study, we carefully designed this Wizard-of-Oz (WoZ) experiment to make participants believe the robot was fully autonomous, ensuring they used natural speech and gestures spontaneously. Upon arrival, participants were led through steps to reinforce this belief, including:

a) clearly informing them that the robot was capable of autonomous operation,

b) showing videos of Baxter performing kitchen tasks, and

c) allowing them to experience Baxter responding to freestyle commands during a practice session.

At the end of the experiment, participants were debriefed on the WoZ design, its purpose, and informed that the robot was remotely tele-operated.}

\subsection{Pilot Studies}
\label{appendix:pilot_studies}
We conducted multiple pilot studies to validate factors that could affect participant behavior to validate independent and dependent control variables as well as the workflow. We experimented with the (a) background noise, (b) perceived robot personality and capability based on the robot's face and name, (c) staging to keep the participants engaged, (d) considerations of the priming effects from practice sessions, (e) \snehesh{WoZ} clues that participants might be able to use to figure out the hidden agenda, and (f) the effect of experiment instructions. These findings informed our experiment design decisions. 

\subsubsection{Background Noise}
\label{appendix:background_noise}
One hypothesis was that background noise can cause people to use more gestures. We considered 3 types of noise recording playback (lawn mower, people talking, and music), but only tested with people talking as background noise as that was the only example people found to be believable and not simulated. We tested at 3 sets of loudness (M (dB) = 58, 63, 70, SD = 10, 13, 15). In our study (N=8), from people's use of speech and gesture and the post-interview, we found that (a) people tune out the background noise, instead of using more gestures or, (b) people wait for gaps of silence or lower level noise in cases of speech or periodic noise, and (c) the noise had to be so loud that none of the speech can be heard at all for them to use gestures instead of speech. For these reasons, we decided not to use background noise as an independent variable.

\subsubsection{Robot Face and Name}
\label{appendix:robot_face_and_name}
To reduce the affect of perceived gender, age, and personality by manipulating facial attributes, we considered the 17 face dimensions based on \cite{Kalegina2018-ki} study to  design the face of the robot to be the most neutral face. The mouth of the robot was removed as not having a mouth did not have significant adverse effect on the neutral perception of the robot.  Having a mouth seemed to give people the idea  that the robot could speak, potentially causes the participant to prefer speech over gesture. For the robot to appear dynamic, friendly, and intelligent, we made the robot blink randomly between 12 and 18 blinks per minute \cite{Takashima2008-tz} with ease-in and ease-out motion profile \cite{Trutoiu2011-se, thomas1995illusion}. We further conducted pilot tests to analyze the head nod motions (velocity and number of nods) and facial expressions for confusion expression. Additionally, we avoided using gender specific pronouns ``he/him" and ``she/her" and referred to the robot as ``the robot" or ``Baxter" which is also the manufacturer given name printed on robot body that tends to be used both as a male and female name \cite{BaxterName_Wikipedia-qz}.

\subsubsection{Staging}
\label{appendix:staging}
As the participants and the robot do \snehesh{not} share the same immersive space, one issue that we observed was when the robot was not directly facing the person and in the middle of a task, the participant loses the context, visibility, and frame of reference. To account for this, we borrow techniques from the 12 principle of animation, specifically \textit{staging} \cite{thomas1995illusion} to gently animate the camera to a view that gives a clean view of the key event that is taking place in the scene. For example, if the robot is pouring oil into the pan, we pick camera angle 2 such that the pan is in the center, with the robot in the background, and oil visible to one side of the screen. One point to note with this addition is when the participant points, we need to log the updated view the participant is looking at for the proper context. For this, we log the camera angle set, robot position, and will need to annotate the pointing direction.

\subsubsection{Practice Session}
\label{appendix:practice_session}
During practice, it is important to make sure that participants are not primed to use one modality versus the other. So steps were taken to design the session with a mixture of related and unrelated commands where both speech and gestures were used to command the robots. If participants used a single modality only, they were encouraged to test out using the other modality. Participants interacted with the robot and asked researchers questions during practice. Once the practice was completed, participants were not allowed to interact with anyone other than the robot even if they had questions or felt stuck as they were told that the experiment was designed for them to experience such scenarios and had to use creative methods make the robot understand what they wanted the robot to do.

\subsubsection{\snehesh{WoZ} Clues}
\label{appendix:woz_clues}
People can be quite intuitive in figuring out the patterns such as key press and mouse click sounds corresponding to robot actions. We experimented with masking the actual clicks and key presses with random ones. However, in the post interview the pilot test participants still seem to be able to figure out that researchers might be controlling the robot. So we created a soft rubber remote control keys that use IR receiver using Arduino micro-controller USB adapter to send keys to the \snehesh{WoZ} UI with virtually no sound that the researcher keeps in their pocket. With this implementation, during the experiment, the researchers made sure when the experiment is being conducted, they do not sit at the control computer and appear to be moving around doing other things appearing busy, staring at their phone seemingly distracted, or looking at the participants showing attention in making sure the system was working without any technical issues. With this implementation, 100\% of the participants believed that the robot was acting on its own and none of the participants suspected the \snehesh{WoZ} setup to be a possibility.

\section{Additional Details on Data Post-processing, Synchronization, and Camera Calibration}
\label{appendix:post_processing_sync_cam_calib}
It is important to clean up the data and make it easy for researchers to use the data. Careful iterative steps were taken to prepare the data to ensure integrity, quality, validity, and fairness. The raw data is processed, annotated, validated, visualized, and curated for downstream analysis and machine learning tasks in the following way.

\textit{i) Multi-camera Calibration}: A standard 12$\times$8 5" checker board was recorded using ROS, and Kalibr package \cite{furgale2013unified} to compute the cameras intrinsic and extrinsic matrix. If the average re-projection error was greater than 1 px., the calibration was repeated.

\textit{ii) Multi-camera Audio-Video Synchronization and Data Compression}: All the data was recorded using ROS bag. These recorded video frames from each cameras tend to have dynamically varying frames per second rates anywhere from 25 fps to 32 fps which makes it difficult to synchronize with sound. For this reason, the audio recording is extracted from a well established audio-video camera such as Apple iPhone camera. A flashing color screen from another computer is placed in the middle of the lab within all the cameras' field of view. A ROS start message is also published to store and identify the starting flag of the session for all other data.
The changing color from red to blue is used to denote the mark of the starting frame and the ROS bag start message time is used for offsetting other messages. The frames are then streamed to a canvas that is 6$\times$ the size of 720p i.e., 2560$\times$2560 where each row is a 720p stereo camera frame. At 33.33ms the latest state of the frame is recorded. The iPhone video is also clipped starting from the blue frames whose sound is then merged with the large canvas video to generate the data. This data is then re-encoded to be compressed using FFMPEG and NVIDIA TITAN X H.264 encoder \cite{ffmpeg-hwaccelIntro-th}. 

\section{Subtasks}
\label{appendix:subtasks}
There are eleven distinct subtasks are featured in NatSGLD dataset (in alphabetical order): 
\[
\begin{array}{|c|c|c|}
\hline
\textit{Add} & \textit{Clean Up} & \textit{Cut} \\
\hline
\textit{Fetch} & \textit{Put On} & \textit{Serve} \\
\hline
\textit{Stir} & \textit{Take Off }& \textit{Transfer} \\
\hline
\textit{Turn Off} & \textit{Turn On} & \\
\hline
\end{array}
\]

\section{Objects}
\label{appendix:objects}
\subsection{Interaction Object Groups}
Objects are categorized into twenty groups, which are:
\begin{itemize}
    \item Food Ingredients: \textit{Carrot, Celery, Pepper, Potato, Salt, Soup, Spices, \snehesh{Oil,} Tomato, Veges}
    \item Cooking Utensils and Tools: \textit{Cutting Board, Knife, Bowl, Spatula, Ladle.}
    \item Cookware: \textit{Lid, Pan, Pot.}
    \item Kitchen Appliances and Fixtures: \textit{Stove, Sink.}
\end{itemize}

\subsection{Object States and Attributes}
\label{appendix:object_states_and_attributes}
Object states and attributes include:
\begin{itemize}
    \item States: \textit{Whole, Cut (Pieces), On, In, Out, Covered, Uncovered, Turned On, Turned Off, and Contains.}
    \item Location and Rotation: Annotated with six degrees of freedom for location and rotation.
    \item Participant Interaction: \snehesh{Gaze as rotation angles}
    and pointing gestures presence are indicated.
\end{itemize}

\section{Multi-Modal Human Task Understanding}
\label{appendix:multi-modal-human-task-understanding}

\subsection{Problem Formulation}
\label{appendix:problem_formulation}

We present the problem of speech-gesture-conditioned task understanding as a translation task that converts a pair of speech and gestures into an LTL formula. This can be seen as a variant of the multi-modal machine translation problem.

In this formulation, the inputs consist of the following components:

\begin{enumerate}
  \item Source Language Vocabulary $\mathcal{V}_{\text{source}}$: This set encompasses words in the source language $\mathcal{L}_{\text{source}}$, which serves as the language in which human instructions are provided.
  \item Gesture Contexts $\mathcal{Z}$: These sequences are derived from human skeleton data and encapsulate diverse gestures made during interactions.
  \item Target LTL Vocabulary $\mathcal{V}_{\text{target}}$: This set encompasses LTL symbols in the target language $\mathcal{L}_{\text{target}}$, which is used to formulate LTL task descriptions.
\end{enumerate}

To solve the speech-gesture-conditioned task understanding problem, the process involves taking a Source Text Sequence $\mathbf{X} = (x_1, x_2, \ldots, x_n)$. This sequence comprises $n$ words from the source language vocabulary, where each word $x_i$ is drawn from $\mathcal{V}_{\text{source}}$. Additionally, there's the Gesture Context $\mathbf{z} = (z_1, z_2, \ldots, z_t)$, which consists of a sequence of $t$ steps of human skeleton data. Each step $z_i$ is a representation of the skeleton's state at that moment, and $\mathbf{z}$ belongs to the set $\mathcal{Z}$.

The desired outcome is a matching target LTL formula $\mathbf{Y} = (y_1, y_2, \ldots, y_m)$. Each element $y_i$ in this sequence belongs to the target LTL vocabulary $\mathcal{V}{\text{target}}$. The solution must ensure that $\mathbf{Y}$ effectively conveys the meaning of the input $\mathbf{X}$ within the context of the target language $\mathcal{L}{\text{target}}$. This task is further complicated by the presence of linguistic and cultural distinctions between the source language $\mathcal{L}{\text{source}}$ and the target language $\mathcal{L}{\text{target}}$.

The learning objective is to minimize the distance between a predicted LTL formula $\hat{Y}=(\hat{y}_1, \hat{y}_2, \ldots, \hat{y}_m)$ and the corresponding target LTL formula $\mathbf{Y} = (y_1, y_2, \ldots, y_m)$:

\begin{equation}\label{loss}
    \mathcal{L}(\mathbf{x,z};\theta)=-\sum_{t=1}^T \mathrm{Cross\-Entropy}(\hat{y}_t(x,z,y_{t-1}), y_t; \theta)
\end{equation}

where $\mathbf{x}$ denotes a text converted from human's speech, $\mathbf{z}$ denotes gesture context features. We use cross-entropy loss to measure the distance between each symbol in predicted and target LTL formulae respectively.

\begin{figure}[ht]
\centering
\includegraphics[width=1.\columnwidth]{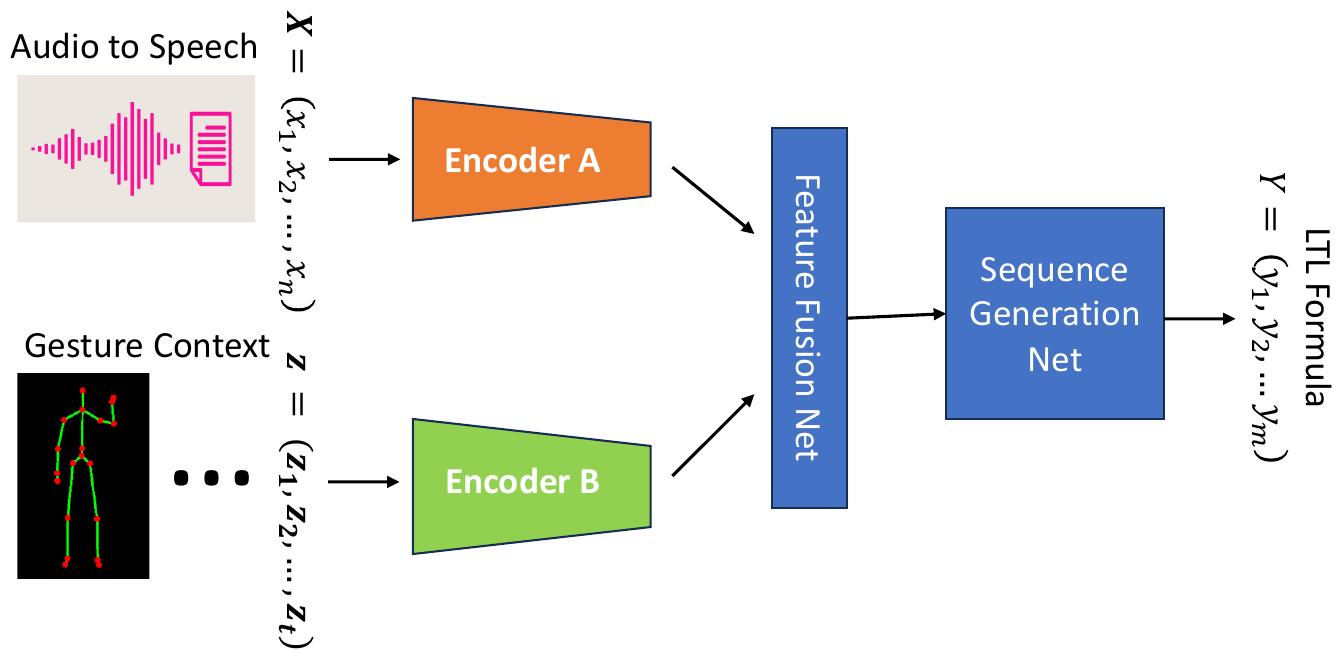}
\caption{The learning framework for translating a pair of speech and gesture data to an LTL formula that can solve multi-modal human task understanding problems.}
\label{fig:lang2ltl_arch}
\end{figure}


\subsection{Approach}
\label{appendix:approach}
Our approach, depicted in Fig. \ref{fig:lang2ltl_arch}, entails a framework for acquiring the ability to forecast an LTL formula from a combination of speech and gesture. The architecture employs a two-stream network that is fed both a word sequence (speech) and a series of skeletal representations (gesture context). This dual input is then processed to yield a sequence of LTL symbols as output. Our training process involves optimizing the network through a cross-entropy loss, in accordance with Eq. \ref{loss}. The Encoder A is implemented by using a pretrained large language model BART \cite{lewis2019bart}. The Encoder B is pretrained by encoding and reconstructing our gesture data.

\subsection{Baselines and Settings}
\label{appedix:baseline_and_settings}
There exists no previous work that specifically addresses the task of learning a mapping from combined natural language sentences and gestures to an LTL formula. The closest work are \cite{wang2021learning,liu2022lang2ltl}, which tackles the challenge of translating natural language sentences into LTL formulas. Given this unique context, we establish two baseline models for comparison, each focusing on using either speech or gestures independently.

\subsubsection{Jaccard Similarity for Logic Formula Evaluation}
\label{appendix:jaccard_similarity_for_ltl_formuala_equivalance}
 Relevant works in logic formula inference have used Jaccard Similarity \cite{li2021explaining,saveri2023towards} as their evaluation metrics. Jaccard Similarity, defined as $J(A, B) = \frac{|A \cap B|}{|A \cup B|}$, measures the similarity between two sets by comparing their intersection $|A \cap B|$ to their union $|A \cup B|$. In evaluating logic formula prediction, Jaccard Similarity quantifies the overlap between the predicted and actual logic formulas, providing a robust assessment by considering the shared elements (true positives) and disregarding non-overlapping elements (false positives and false negatives).

\subsubsection*{Spot Score for LTL Formula Equivalence}
\label{appendix:spot_score_for_ltl_formuala_equivalance}
In addition to Jaccard Similarity, we introduce a novel accuracy metric that tests the equivalence between two LTL formulas using the Spot library \cite{duret2016spot, duret2022spot}. The Spot library functions by converting input \(f\) and \(g\) (as well as their negations) into four automata: \(A_f\), \(A_{\neg f}\), \(A_g\), and \(A_{\neg g}\). It then ensures that both \(A_f \otimes A_{\neg g}\) and \(A_g \otimes A_{\neg f}\) are empty, validating formula equivalence. This metric, previously utilized in \cite{liu2022lang2ltl}, provides a precise measure for evaluating LTL formula predictions.

To determine if predicted and ground-truth formulas \(f_1\) and \(f_2\) are equivalent, we employ the \texttt{spot.are\_equivalent()} function, illustrated in the following link\footnote{\url{https://spot.lre.epita.fr/tut04.html}}. We provide a sample Python code snippet for reference:

\begin{verbatim}
import spot

# Define two Spot formulas
f1 = spot.formula("GF(a & Xb)")
f2 = spot.formula("G(Fa & Fb)")

# Check if the formulas are equivalent
if spot.are_equivalent(f1, f2):
    print("Equivalent.")
else:
    print("Not equivalent.")
\end{verbatim}

Here, \texttt{f1} and \texttt{f2} represent the Spot formulas under comparison. The \texttt{spot.are\_equivalent()} function returns \texttt{true} if the formulas are equivalent, and \texttt{false} otherwise. By assessing the equivalence of all pairs of predicted and annotated LTL formulas, we calculate the Spot Score:

\begin{equation}\label{eq:spot-score}
    \text{Spot-Score} = \frac{\sum_{i=1}^N \texttt{spot.are\_equivalent}(f_i, g_i)}{N}
\end{equation}

\subsection{Results and Analysis}
\label{appendix:results_and_analysis}

We conduct a comprehensive assessment, combining quantitative and qualitative analyses, to underscore the significance of incorporating speech and gestures in Multi-Modal Human Task Understanding tasks. In this endeavor, we introduce a comparative analysis between two state-of-the-art large language models (LLMs): T5, developed by Google, and BART, created by Meta. This comparison aims to evaluate their efficacy in predicting Linear Temporal Logic (LTL) formulas from inputs comprising either speech alone or a combination of speech and gestures.

Tables \ref{tab:table1} and \ref{tab:table2} showcase the performance metrics for these models, illustrating the distinct benefits of leveraging both speech and gestures for a deeper understanding of tasks articulated by humans. Our findings indicate that integrating speech with gestures not only enhances model performance but also significantly outperforms the results obtained by relying on a single modality.

\begin{table}[h] \label{table:BART_spot}
    \centering
    \begin{tabular}{|l|c|r|}
        \hline
        \textbf{Model (using BART \cite{lewis2019bart})} & \textbf{Jaq Sim$\uparrow$} & \textbf{Spot Score$\uparrow$} \\
        \hline
        Speech Only & 0.934 & 0.434 
        \\
        \hline
        Gestures Only & 0.922 & 0.299 
        \\
        \hline
        Speech + Gestures & \textbf{0.944} & \textbf{0.588} 
        \\
        \hline
    \end{tabular}
    \caption{Performance Comparison: Jaccard Similarity and Spot Score results for LTL formula prediction using BART}

    \label{tab:table1}
\end{table}

\begin{table}[h]
    \centering
    \begin{tabular}{|l|c|r|}
        \hline
        \textbf{Model (using T5 \cite{raffel2020exploring})} & \textbf{Jaq Sim$\uparrow$} & \textbf{Spot Score$\uparrow$} \\
        \hline
        Speech Only & 0.917 & 0.299 
        \\
        \hline
        Gestures Only & 0.948 & 0.244 
        \\
        \hline
        Speech + Gestures & \textbf{0.961}  & \textbf{0.507}
        \\
        \hline
    \end{tabular}
    \caption{Performance Comparison: Jaccard Similarity and Spot Score outcomes for LTL formula prediction using T5}

    \label{tab:table2}
\end{table}

Furthermore, the juxtaposition of T5 and BART models in our analysis yields additional insights into the nuanced ways different Language Model architectures process and interpret multi-modal data. The comparative results presented in Tables \ref{tab:table1} and \ref{tab:table2} affirm the premise that, although speech is the predominant mode of communication, the inclusion of gestures provides indispensable contextual support, enhancing the interaction model's comprehension capabilities.

\section{Additional Acknowledgments}
\label{appendix:additional_acknowledgements}
We thank our peers, faculties, staff, and research assistants from the Perception and Robotics Group (PRG), the University of Maryland Computer Science department (UMD CS), and the University of Maryland Institute for Advanced Computer Studies (UMIACS) for their continued support, their valuable discussions and feedback. 

Special thanks to Dr. Nirat Saini, Dr. Virginia Choi, Dr. Chethan Parameshwara, Dr. Nitin Sanket, and Dr. Chahat Deep Singh. We want to thank and recognize the contributions of Aavash Thapa, Sushant Tamrakar, Jordan Woo, Noah Bathras, Zaryab Bhatti, Youming Zhang, Jiawei Shi, Zhuoni Jie, Tianpei Gu, Nathaneal Brain, Jiejun Zhang, Daniel Arthur, Shaurya Srivastava, and Steve Clausen. Without their contributions and support, this work would not have been possible. Finally, a special thanks to Dr. Jeremy Marvel of the National Institute of Standards and Technology (NIST) for his support and advice.

\end{appendices}

\end{document}